\documentclass[10pt,twocolumn,letterpaper]{article}

\usepackage[pagenumbers]{cvpr} 

\usepackage{graphicx}
\usepackage{amsmath}
\usepackage{amssymb}
\usepackage{booktabs}
\usepackage{algorithm}
\usepackage{algpseudocode}
\usepackage{mathrsfs}

\usepackage{multirow}
\usepackage{float}
\graphicspath{{./imgs/}}

\usepackage[pagebackref,breaklinks,colorlinks]{hyperref}

\usepackage[capitalize]{cleveref}
\usepackage{comment}
\crefname{section}{Sec.}{Secs.}
\Crefname{section}{Section}{Sections}
\Crefname{table}{Table}{Tables}
\crefname{table}{Tab.}{Tabs.}

\def\cvprPaperID{11004} 
\def\confName{CVPR}
\def\confYear{2023}

\begin{document}

\title{StyLess: Boosting the Transferability of Adversarial Examples}

\author{Kaisheng Liang \qquad Bin Xiao\thanks{Corresponding author.}\\
The Hong Kong Polytechnic University\\
{\tt\small \{cskliang, csbxiao\}@comp.polyu.edu.hk}
}
\maketitle

\begin{abstract}
Adversarial attacks can mislead deep neural networks (DNNs) by adding imperceptible perturbations to benign examples.
The attack transferability enables adversarial examples
to attack black-box DNNs with unknown architectures or parameters,
which poses threats to many real-world applications.
We find that existing transferable attacks do not distinguish between style and content features during optimization,
limiting their attack transferability.
To improve attack transferability, we propose a novel attack method called style-less perturbation (StyLess).
Specifically, instead of using a vanilla network as the surrogate model, we advocate using stylized networks,
which encode different style features by perturbing an adaptive instance normalization.
Our method can prevent adversarial examples from using non-robust style features
and help generate transferable perturbations.
Comprehensive experiments show that our method can significantly improve the transferability of adversarial examples.
Furthermore, our approach is generic and can outperform state-of-the-art transferable attacks when combined with
other attack techniques.
\footnote{Our code is available at https://github.com/uhiu/StyLess}
\end{abstract}

\section{Introduction}
Deep neural networks (DNNs) \cite{krizhevsky2012imagenet,He_2016_CVPR} are currently effective methods for solving various challenging tasks such as computer vision, and natural language processing.
Although DNNs have amazing accuracy, especially for computer vision tasks such as image classification, they are also known to be vulnerable to adversarial examples \cite{SzegedyZSBEGF13,GoodfellowSS14}.
Adversarial examples are malicious images obtained by adding imperceptible perturbations to benign images.
Notably, the transferability of adversarial examples is an intriguing phenomenon,
which refers to the property that the same adversarial example can successfully attack different black-box DNNs \cite{papernot2016transferability,LiuCLS17,DongLPS0HL18,XieZZBWRY19}.

It has been observed that image style can be decoupled from image content, and style transfer techniques allow us to generate stylized images based
on arbitrary style images \cite{huang2017arbitrary}.
Image style refers to the unique visual characteristics of an image, including its colors, textures, and lighting.
For instance, two photos of the same object taken by different photographers can have very different styles.
Robust DNNs should rely more on content features of data than style features.
This inspired us to improve attack transferability from the perspective of avoiding non-robust features.
We believe that style features of DNNs are non-robust for building transferable attacks. However, existing attacks do
not distinguish between the surrogate models' style and content features, which may reduce attack
transferability.

\begin{figure}
    \setlength{\abovecaptionskip}{0.2cm}
\centering
\includegraphics[width=0.47\textwidth]{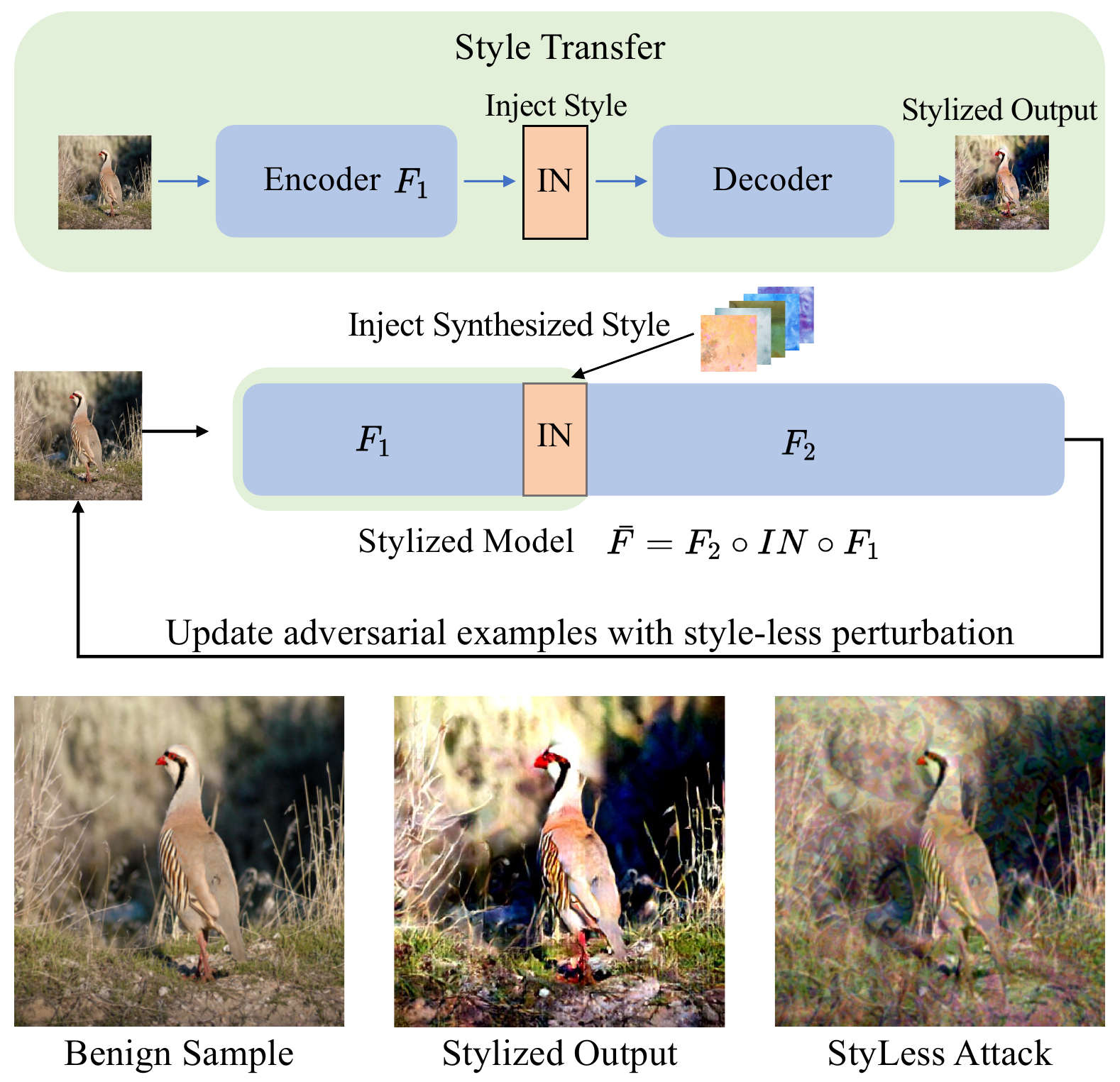}
\caption{
An overview of our StyLess attack.
We create stylized model $\bar F$ by injecting synthesized style features into the surrogate model ($F=F_2 \circ F_1$) using an adaptive IN layer. 
StyLess reduces the use of non-robust style features in the vanilla surrogate model $F$, ultimately improving attack transferability.
}
\label{fig:model_sip}
\vspace{-3ex}
\end{figure}

We propose using stylized surrogate models to control style features, which can significantly improve transferability.
We refer to the original surrogate model as the ``vanilla model."
The proposed stylized model is created by adding an adaptive instance normalization (IN) layer to the vanilla model.
By adjusting the parameters of the inserted IN layer, we can easily transform the style features of the surrogate models.
To compare the stylized and vanilla surrogate models, we analyzed their network losses during optimization. 
Surprisingly, we found that the adversarial loss of the vanilla model increased much faster than those of the stylized models, resulting in a widening loss gap.
This phenomenon reveals that as the attack iteration progresses, current attack methods only focus on maximizing the loss of the vanilla model,
leading to increased use of the style features of the vanilla surrogate model.
However, we believe that style features are non-robust for transferable attacks, and relying too much on them may reduce attack transferability.
To enhance transferability, we aim to limit the use of non-robust style features and close the loss gap.

Based on the above findings, we propose a novel method called StyLess to improve the transferability of adversarial examples.
Our method uses multiple synthesized style features to compete with the original style features during the iterative optimization of attack.
The process is illustrated in Figure~\ref{fig:model_sip}.
We encode various synthesized style features into a surrogate model via an IN layer to achieve stylized surrogate models.
Instead of using only the vanilla surrogate model, we use the gradients of both the stylized surrogate models and the
vanilla one to update adversarial examples.
The front part of the surrogate model works as a style encoder, and the IN layer simulates synthesized style features.
Although we can use a decoder to explicitly generate the final stylized samples, it is unnecessary for the proposed attack method. %
Experimental results demonstrate that StyLess can enhance the transferability of state-of-the-art adversarial attacks on both unsecured and secured black-box DNNs.
Our main contributions are summarized as follows:
\vspace{-1ex}
\begin{itemize}
\setlength{\itemsep}{1pt}
\setlength{\parskip}{0pt}
\item We introduce a novel perspective for interpreting attack transferability: the original style features may hinder transferability.
We verify that current iterative attacks increasingly use the style features of the surrogate model during the optimization process.
\item We propose a novel attack called StyLess to enhance transferability by minimizing the use of original style features.
To achieve this, we insert an IN layer to create stylized surrogate models and use gradients from both stylized and vanilla models.
\item We conducted comprehensive experiments on various black-box DNNs to demonstrate that StyLess can significantly
improve attack transferability. Furthermore, we show that StyLess is a generic approach that can be combined with existing attack techniques.
\end{itemize}

\section{Related Work}
\label{sec:related}
\textbf{Adversarial Attacks.} Adversarial attacks reveal the vulnerability of current DNNs \cite{SzegedyZSBEGF13}.
The classic adversarial attack methods are gradient-based, such as FGSM \cite{GoodfellowSS14}
and I-FGSM \cite{KurakinGB17a}.
C\&W \cite{carlini2017towards} considers optimizing the distance between adversarial examples and benign samples, and
proposed optimization-based attacks.
Adversarial attacks can also be performed in the physical world \cite{eykholt2018robust,sharif2016accessorize}.
As for defending against adversarial examples, adversarial training is a popular defense method that
uses adversarial examples as extra training data to improve robustness \cite{MadryMSTV18}.

\textbf{Increasing Attack Transferability.}
An intriguing property of adversarial attacks is the transferability.
Ensemble-based attack \cite{LiuCLS17} uses multiple surrogate networks instead of one network.
Ghost networks \cite{LiBZXZY20} generates different surrogate networks by perturbing skip connection and dropout layers.
Optimization methods, such as MI \cite{DongLPS0HL18}, uses a momentum-based optimization,
while VT \cite{Wang021} introduces gradient variance to control the stability of the localized gradients.
RAP \cite{qin2022boosting} generates adversarial examples located in a flat loss region.
Data augmentation methods, such as DI \cite{XieZZBWRY19}, uses image transformation like resizing and padding, while TI \cite{DongPSZ19} considers translating image pixels.
SI \cite{LinS00H20} calculates gradients with the help of several scaled benign samples.
Admix \cite{Wang_2021_admix} calculates iterative gradients by mixing the benign images with randomly sampled images.

Various network architectures and features exhibit different relationships with adversarial attacks.
DNNs’ linearity is believed to cause adversarial vulnerability \cite{GoodfellowSS14}, and LinBP \cite{GuoLC20} skips the nonlinear activation during the backpropagation.
SGM \cite{Wu0X0M20} uses more gradients through skip connections in residual networks.
To better leverage the intermediate layers, one can train auxiliary classifiers based on feature spaces \cite{InkawhichLCC20,InkawhichLWICC20},
maximize the distance between natural images and their adversarial examples in feature spaces \cite{ZhouHCTHGY18},
or fine-tune the existing adversarial examples in intermediate layer level by ILA \cite{HuangKGHBL19,LiGC20}.

\textbf{Style Transfer and Instance Normalization.}
Style transfer can change the style of an image to match the style of another one
 \cite{gatys2016image,johnson2016perceptual,dumoulin2016learned}.
Fast feedforward networks can perform stylization with arbitrary styles in a single forward pass
\cite{huang2017arbitrary, li2017universal}.
Interestingly, style transfer has a wide range of applications.
AdvCam \cite{Duan_2020_CVPR} uses natural styles to hide non-$L_p$ restricted perturbations.
FSA \cite{xu2021towards} generates natural-looking adversarial examples by using optimized style changes.
Style transfer has also been used to improve network robustness by exploring additional feature information \cite{naseer2022stylized}.
Latent style transformations can detect adversarial attacks \cite{wang2022adversarial}.
AMT-GAN \cite{Hu_2022_CVPR} proposes an adversarial makeup transfer to protect facial privacy by preserving stronger black-box transferability.

The family of instance normalization (IN) including batch normalization \cite{IoffeS15},
layer normalization \cite{BaKH16}, instance normalization \cite{UlyanovVL16}, and group normalization \cite{WuH18}.
Normalizations are mainly used to reduce the covariate shift, and speed model training.
Recently, normalizations have been found to be related to robustness.
It has been shown that batch normalization makes DNNs use more non-robust but useful features \cite{benz2021batch,IlyasSTETM19}.
AdvBN proposed adding an extra batch normalization into network training to increase training loss adversarially, which enables the network to resist various domain shifts \cite{shu2021encoding}.
Adjusting batch normalization statistics such as the running mean and variance in the inference phase,
which are estimated during training, improves robustness and defense common corruption \cite{SchneiderRE0BB20,BenzZKK21}.

Among existing style-based attacks,
FSA \cite{xu2021towards} differentiates style features and content features,
which is similar to our method.
However, there are three significant differences between FSA and our approach:
1)
FSA proposes to hide adversarial perturbations in the optimized style,
while we avoid relying on any style.
2)
FSA aims at enhancing the natural looking of non-$\ell_p$ restricted attacks,
while we focus on the transferability of $\ell_p$ restricted adversarial examples.
3)
Both FSA and our work are inspired by AdaIN \cite{huang2017arbitrary}, but we use IN layer differently.
FSA perturbs the IN layer to search malicious styles and requires a decoder.
But we use randomized IN layers to augment attacks and don't need to train a decoder.

\section{Methodology}\label{sec:method}
\subsection{Threat Model}
\label{method:bg}
\textbf{Attack objective.}
Given a benign image $x$ with label $y$,
transfer-based attacks aim to generate an adversarial perturbation based on a white-box surrogate network $F$.
The general attack objective can be formulated as follows:
\begin{equation}\label{eq:pre_att_obj}
    \max_{\delta} \mathcal{L}(F(x+\delta), y) \quad \text{s.t.} \quad \lVert \delta \rVert \leq \epsilon,
\end{equation}
where $\mathcal{L}$ denotes the adversarial loss,
$\delta$ is the  adversarial perturbation,
and $\epsilon$ is the maximum perturbation size.

A popular framework to solve the above problem
is iterative fast gradient sign method (I-FGSM) \cite{GoodfellowSS14,KurakinGB17a}:
\begin{equation}\label{eq:ifgsm}
 {x}^{t+1}_{adv}={x}^{t}_{adv}+\alpha \cdot \operatorname{sign}\left(\nabla_{{x
 }} \mathcal{L}\left(F({x}^{t}_{a d v}), y \right)\right),
\end{equation}
where $\alpha$ is the learning rate,
and a clip function will be used on $x^{t+1}_{adv}$ to ensure
$\lVert {x}^{t+1}_{adv}-x \rVert \leq \epsilon$.

\textbf{Attacker capability.}
We follow the same setting in previous work that
attackers have a surrogate model and some test samples,
but cannot access target models, and don't know network architectures,
training data, or defense strategies.
It should be noted that our method doesn't require any additional datasets.
Our approach involves style features, which can be extracted from an arbitrary image or synthesized without any style image.

\textbf{Transferable attacks as black-box attacks.}
Transferable attacks use the surrogate model $F$ to create adversarial examples that can fool unseen target models.
In this way, these attacks can be viewed as black-box attacks.

\subsection{Motivation}
Existing transferable attacks often rely on the gradient of the adversarial loss function $\mathcal{L}$ (Equation~\ref{eq:ifgsm})
without considering the impact of different components of $\mathcal{L}$.
However, these approaches have limitations because transferable attacks should minimize the use of non-robust features of the surrogate model.
Interestingly, for image classification task, style features of images are typically less robust
than content features.
Based on this observation, we propose to enhance attack transferability by explicitly
reducing the use of style features of the surrogate model within the loss $\mathcal{L}$.

Our key idea is simulating various surrogate models without the style features of the given vanilla surrogate model.
We discovered that inserting an IN layer into the vanilla surrogate model enables us to create new surrogate models
that we refer to as \textit{stylized surrogate models}.
Sometimes we omit the word ``surrogate."
Stylized models can explicitly manipulate style features without compromising model accuracy.
Recall that our goal is to construct adversarial examples that can mislead unseen target models, which should include these stylized models.
However, existing methods, such as MI and I, only focus on maximizing the loss of the vanilla model.

\begin{figure}[htb]
\centering
\begin{subfigure}{0.44\textwidth}
\includegraphics[width=\linewidth]{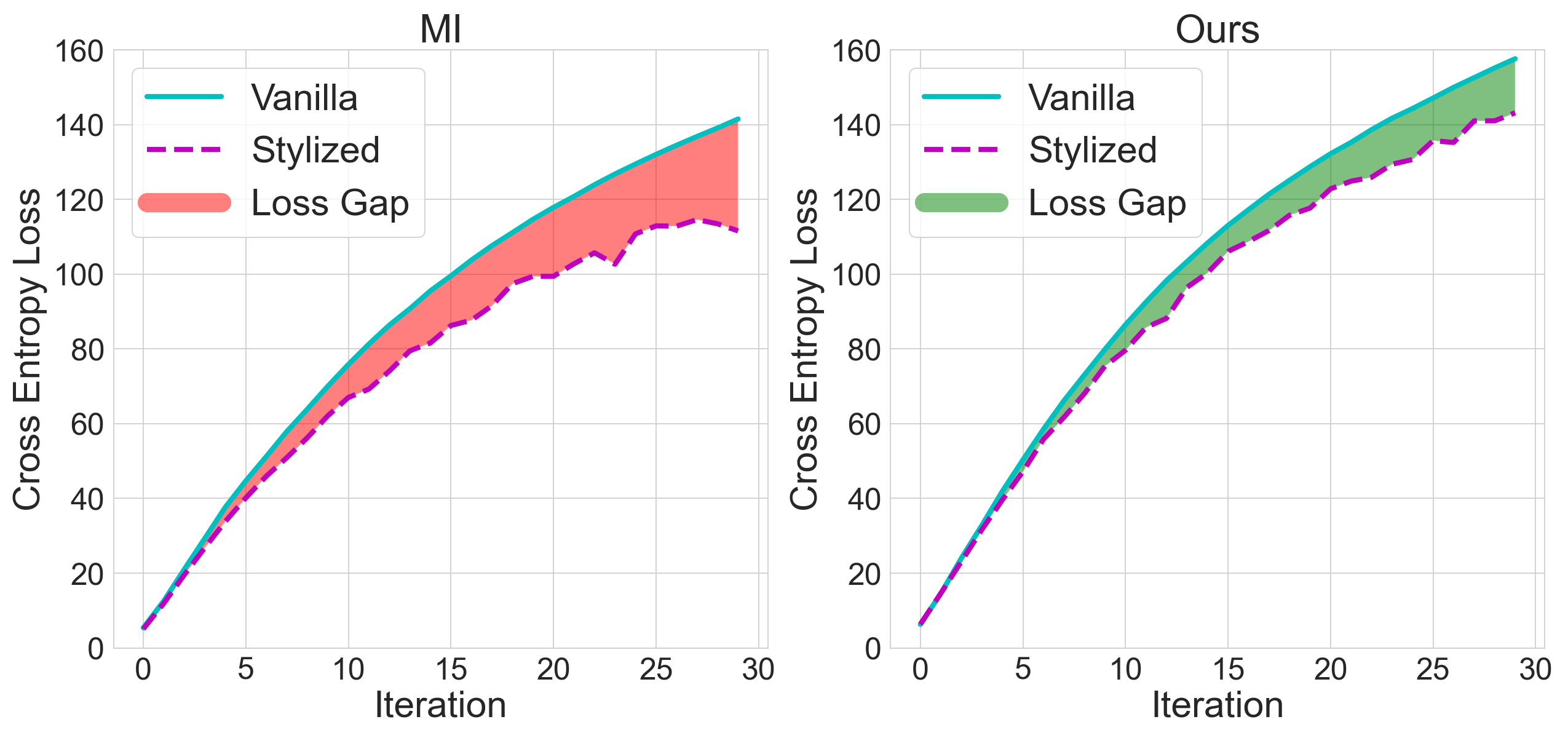}
\caption{Baseline attack MI vs ours MI+StyLess}
\end{subfigure}
\begin{subfigure}{0.44\textwidth}
\includegraphics[width=\linewidth]{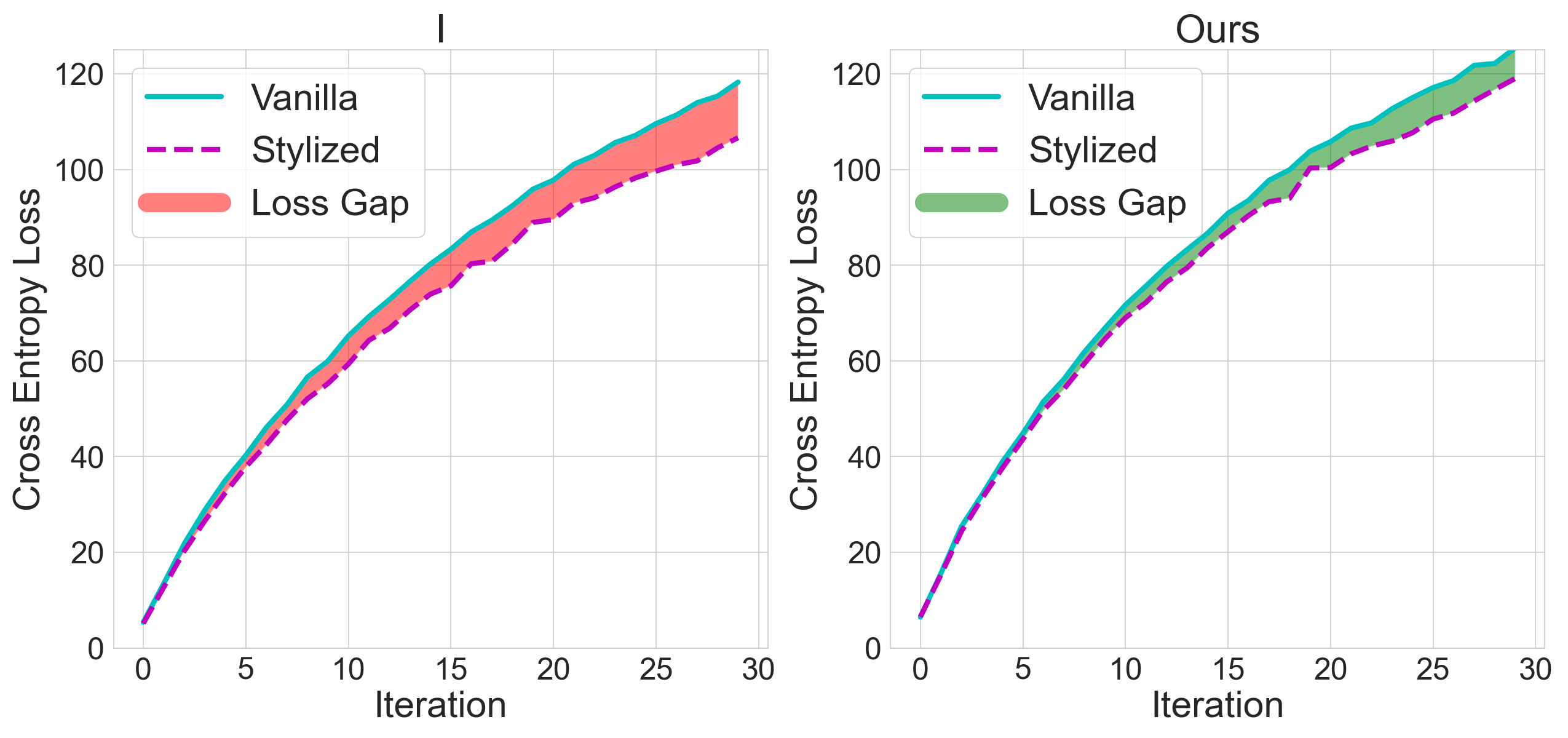}
\caption{Baseline attack I(-FGSM) vs ours I+StyLess}
\end{subfigure}
\caption{Illustration of the loss gap between vanilla network and stylized network,
with RN50 as surrogate model.
The greater the loss, the better the attack performance.
The widening loss gap in the baseline means that attack performance on stylized models is lagging behind.
The proposed StyLess can close the gap.}
\label{fig:gap_stylized}
\vspace{-1ex}
\end{figure}

Figure~\ref{fig:gap_stylized} indicates that MI and I(-FGSM) have limited attack transferability on stylized models
since the vanilla model’s adversarial loss increases much faster than stylized models', resulting in a widening loss gap.
In the following sections, we will demonstrate how our method addresses this issue by maximizing the loss of both the
stylized and vanilla models, which significantly improves transferability.

\subsection{Stylized Surrogate Models}
\label{method:CoSG}
This section will give the definition of our stylized surrogate models, which encode various style features by
inserting an IN layer.
Then we will analyze the stylized loss gap ($\Delta\mathcal{L}$) between the vanilla and stylized surrogate models.
Specifically, we will illustrate the increasing $\Delta\mathcal{L}$ limits transferability and our idea to decrease $\Delta\mathcal{L}$.
\vspace{-2ex}
\subsubsection{Encoding Styles by Stylized Models}
Given a classifier $F=F_2 \circ F_1$  as the surrogate model,
we define a stylized surrogate model as
\begin{equation}\label{eq:stylized_f}
    \bar F_{x_s} 
    = F_2 \circ \operatorname{IN}_{x_s} \circ F_1,
\end{equation}
where $x_s$ is a style input, $\operatorname{IN}_{x_s}$ is an IN layer instantiated by $x_s$.
In general, an IN layer is defined as
\begin{equation}
\operatorname{IN}(x; \mu, \sigma) 
=\sigma \cdot \left(\frac{x-\mu(x)}{\sigma(x)}\right)+\mu,\label{eq:IN}
\end{equation}
where $\mu \text{ and } \sigma$ are the network parameters of layer IN,
and $\mu(x) \text{ and } \sigma(x)$ are the mean and variance of input $x$.
According to adaptive instance normalization (AdaIN) \cite{huang2017arbitrary},
to stylize an input $x$ with a given style input $x_s$, we only need to instantiate the IN as
\begin{equation}
\operatorname{IN}_{x_s}(x) = \operatorname{IN}(x)
\left|_{\mu=\mu(x_s), \sigma=\sigma(x_s)}\right. \label{eq:IN_style}
\end{equation}

Our stylized model $\bar F_{x_s}$ has encoded style features.
Based on AdaIN, the $F_1$  from the classifier $F=F_2 \circ F_1$ works as an encoder for style transfer,
and given a style input $x_s$, we can get a stylized image as
$\bar x = D \circ \operatorname{IN}_{x_s} \circ F_1(x)$,
where $D$ denotes a decoder.
Thus, $\operatorname{IN}_{x_s} \circ F_1(x)$ has encoded the style features of $x_s$
to $\bar F_{x_s}$.

\vspace{-2ex}
\subsubsection{Stylized Loss Gap Limits Transferability}
To validate the performance of adversarial attacks on these stylized
surrogate models, we define a \textit{stylized loss gap} as
\begin{equation}
\Delta\mathcal{L}=
\mathbb{E}_{x_s \in \mathcal{D}}[
{\mathcal{L}} (F(x), y) - {\mathcal{L}} (\bar F_{x_s}(x), y)],
\end{equation}
where $F$ and $\bar F_{x_s}$ are vanilla and stylized model, respectively;
$x_s$ is style input.

An increasing loss gap $\Delta\mathcal{L}$ can be observed in Figure~\ref{fig:gap_stylized}
which limits attack transferability, as explained below.
Hypothetically, let's say we can decouple style-dependent loss from content-dependent loss in $\mathcal{L}$ as follows:
\begin{equation}\label{eq:decoupled}
    \begin{aligned}
        \mathcal{L}       &= \mathcal{L}^c + \mathcal{L}_{x}^s,\\
        \bar{\mathcal{L}} &= \mathcal{L}^c + \mathcal{L}_{x_s}^s,
    \end{aligned}
\end{equation}
where $\mathcal{L}$ is the vanilla loss, and $\bar{\mathcal{L}}$ is the stylized one;
$\mathcal{L}^c$ is the common content-dependent loss;
$\mathcal{L}_{x}^s$ is input sample $x$-specific style-dependent loss,
while $\mathcal{L}_{x_s}^s$ is style sample
$x_s$-specific style-dependent loss.
In this case, $\Delta\mathcal{L}=\mathcal{L}^s_x-\mathcal{L}^s_{x_s}$.

In other words, the non-robust features are related to $\mathcal{L}_x^s$ and $\mathcal{L}_{x_s}^s$, while ${\mathcal{L}}_c$ is supposed to be shared by other unseen DNNs.
MI and I-(FGSM) have increasing gaps in Figure~\ref{fig:gap_stylized} means that attackers gradually focus on
optimizing the $\mathcal{L}_x^s$ part, which only belonged to the vanilla surrogate model $F$.
Therefore, the loss gap limits the transferability of adversarial examples in unseen stylized models.

To decrease $\Delta\mathcal{L}$ and boost transferability,
we consider involving $\mathcal{L}_{x_s}^s$ as a competitor of
$\mathcal{L}_{x}^s$ in optimization process to suppress the growth of $\mathcal{L}_{x}^s$.
In general, we can assume that all these losses in Equation~\ref{eq:decoupled} are non-negative, and they satisfy:
$\mathcal{L}^c \gg \mathcal{L}_{x}^s, \ \mathcal{L}^c \gg \mathcal{L}_{x_s}^s$.
Also, there is a upper bound $B$ for these losses: $\mathcal{L}<B, \ \bar{\mathcal{L}} < B$,
as the adversarial perturbations are required to smaller than a given $\epsilon$.
If we only maximize the vanilla loss $\mathcal{L}$, both $\mathcal{L}^c \text{ and }
\mathcal{L}_{x}^s$ are likely to be increased.
To this end, we propose to maximize
$\mathbb{E}_{x_s \in \mathcal{D}} \mathcal{L}_{x_s}^s +
\mathcal{L}_{x}^s + \mathcal{L}^c$,
which involves multiple style-dependent losses
$\mathcal{L}_{x_s}^s (x_s \in \mathcal{D})$ to compete with
$\mathcal{L}_{x}^s$ and leads to a decrease of $\Delta\mathcal{L}$.

\subsection{Proposed Style-Less Perturbations (StyLess)}
\label{method:sip}
Based on the above analysis,
we propose style-less perturbations (StyLess) method to increase attack transferability by optimizing stylized loss and vanilla loss together:
\begin{equation}\label{eq:obj_StyLess}
\max_{\delta} \mathbb{E}_{x_s \in \mathcal{D}}
{\mathcal{L}} (\bar F_{x_s}(x+\delta), y)
+{\mathcal{L}} (F(x+\delta), y).
\end{equation}

The key to generate multiple stylized models $\bar F_{x_s}$ 
is synthesizing style statistics $\mu, \sigma$ for Equation~\ref{eq:IN_style} to obtain parameterized IN layers.
We propose using scaling and interpolation to simulate multiple style features, formulated as
\begin{equation}
    \label{eq:sip_factor}
    \begin{aligned}
        \mu    &= \beta  (\lambda \mu_x    +(1-\lambda)\mu_s), \\
        \sigma &= \gamma (\lambda \sigma_x +(1-\lambda)\sigma_s),
    \end{aligned}
\end{equation}
where $\mu_x, \sigma_x$ is the mean and variance of $F_1(x)$, relating to the benign content input $x$,
while $\mu_s, \sigma_s$ are for style input $x_s$ similarly.
$x_s$ is an arbitrary image.
$\lambda$ is a scalar that controls the interpolation of
two styles. 
$\beta$ and $\gamma$ are $c$-dimensional vectors that scale the synthesized style,
where $c$ refers to the number of channels in the style feature.

We summarize the proposed StyLess attack in Algorithm~\ref{alg:sip}.
To obtain the parameters of the IN layer without training,
we select a random image as $x_s$, and generate a pair of $\mu$ and $\sigma$ using Equation~\ref{eq:sip_factor}.
With $\mu$ and $\sigma$, we can create a stylized model $\bar F_{x_s}(x)$ using Equation~\ref{eq:stylized_f} and ~\ref{eq:IN_style}.
$\bar F_{x_s}(x)$ need to be equal to $F(x)$ or label y, otherwise we will regenerate $\mu$ and $\sigma$ by altering $\beta, \gamma \text{ and } \lambda$.
In each iteration, we use $N$ stylized models to update adversarial examples based on the proposed objective function in Equation~\ref{eq:obj_StyLess}.
\begin{algorithm}[h]
    \algnewcommand\algorithmicinput{\textbf{Input:}}
    \algnewcommand\Input{\item[\algorithmicinput]}
    \algnewcommand\algorithmicoutput{\textbf{Output:}}
    \algnewcommand\Output{\item[\algorithmicoutput]}
    \caption{Style-Less Perturbations (StyLess) Algorithm}
    \label{alg:sip}
	\begin{algorithmic}[1]
		\Input Surrogate model $F$, 
               benign example $x$, 
               iteration number $T$, maximum perturbation $\epsilon$,
               data augmentation $\phi(\cdot)$, decay factor $\eta$.
               Scale factors $\beta, \gamma$ and interpolation factor $\lambda$.
               The number of stylized models $N$.
        \Output An adversarial example $x_{adv}$
        \State $x_{adv}=x, g_0=0, \alpha = \epsilon/2.$
		\For{$t = 0 \rightarrow T-1$}:
            \State Augment input $x_{adv}=\phi(x_{adv})$
            \State Obtain gradient $\tilde g_{t+1}$ with respect to $x_{adv}$ using $F$
            \Repeat
            \State Synthesize a style statistic by Equation~\ref{eq:sip_factor}
            \State Obtain a stylized model $\bar F_{x_s}$ by Equation~\ref{eq:stylized_f}
            \State Get gradient $\tilde g$ with respect to $x_{adv}$ using $\bar F_{x_s}$
            \State Update $\tilde g_{t+1} = \tilde g_{t+1} + \tilde g$
            \Until obtain $N$ stylized models
            \State Calculate momentum $g_{t+1} = \eta \cdot g_{t} + \tilde g_{t+1}/\|\tilde g_{t+1}\|_1$
            \State Update example
		        $x_{adv} = x_{adv} + \alpha \cdot \text{sign}(g_{t+1})$
		\EndFor
        \State \Return $x_{adv}$.
	\end{algorithmic}
\end{algorithm}

\section{Experiments}\label{sec:exp}

\subsection{Experimental Setup}\label{exp:setup}

\textbf{Dataset.}
We use ImageNet \cite{russakovsky2015imagenet} for experiments.
Specifically, we use 1000 images which are randomly selected from ImageNet validation set by Wang et al. \cite{Wang021}.
Similar dataset settings have been widely used in previous work
\cite{Wang021,DongLPS0HL18,LinS00H20,XieZZBWRY19,Wu0X0M20,GuoLC20}.
These images include all categories; almost all are correctly classified by the target DNNs.

\textbf{Models.}
We evaluate the generated adversarial examples on different black-box DNNs, including both unsecured and secured models.
Unsecured models are trained on ImageNet using traditional methods, while secured models are based on adversarial training.
The unsecured models include VGG19 \cite{SimonyanZ14a}, AlexNet \cite{krizhevsky2012imagenet}, ResNet50 (RN50) \cite{He_2016_CVPR},
WideResNet101 (WRN101) \cite{Zagoruyko2016WRN}, DenseNet121 (DN121) \cite{huang2017densely}, InceptionV3 (IncV3) \cite{SzegedyVISW16},
MnasNet \cite{tan2019mnasnet}, MobileNetV2 (MNv2) \cite{sandler2018mobilenetv2}, ShuffleNetV2 (SNv2) \cite{ma2018shufflenet} and ViT \cite{dosovitskiy2020vit}.
Their pre-training parameters are obtained from PyTorch official.
The secured DNNs are IncV3$_{\text{ens3}}$ (ensemble of 3 InceptionV3 networks),
IncV3$_{\text{ens4}}$ (ensemble of 4 InceptionV3 networks) and IncResV2$_{\text{ens3}}$ (ensemble of 3 IncResV2
networks).
These models were adversarially trained and widely used in previous work \cite{DongLPS0HL18,Wu0X0M20,Wang021,LiuCLS17}.
As for the surrogate models, we use VGG19 \cite{SimonyanZ14a}, RN50 \cite{He_2016_CVPR},
WRN101 \cite{Zagoruyko2016WRN} and DN121 \cite{huang2017densely}.

\textbf{Implementation Details.}
We use I-FGSM \cite{GoodfellowSS14} as the initial baseline.
Unless otherwise specified, the attacks are untargeted and $l_{\infty}$-restricted.
The maximum perturbation size is set to $\epsilon = 16/255$.
We compare StyLess primarily with six transferable attacks: MI \cite{DongLPS0HL18}, DI \cite{XieZZBWRY19}, TI \cite{DongPSZ19}, SI \cite{LinS00H20}, Admix (AI)\cite{Wang_2021_admix},
and an ensemble-based approach \cite{LiuCLS17}.
We set the optimization step size to $\alpha=\epsilon/2$, and the number of iterations to $T=50$. The momentum decay in MI is $\mu=1$.
For DI, SI and AI, we follow the official settings described in the corresponding papers.
For StyLess,
we simulate 10 stylized models in each iteration, denoted by $N=10$.
To generate a stylized model for a given $x$,
we randomly sample $\lambda$ from $[0,0.2]$,
and $\beta, \gamma$ from $[0,2]$, and ensure that $\bar F_{x_s}(x)$ is equal to $F(x)$ or the real label.
The IN layer is inserted after the first bottleneck block for RN50 and WRN101, and after the first dense block for DN121.

\subsection{Attacking Unsecured Models}
We compare StyLess with other attacks on various unsecured DNNs using three surrogate models.
Table~\ref{table:full-unsecured} shows that StyLess is a powerful and generic method that can be combined with existing attack methods to further improve attack transferability.
Specifically, we compare StyLess with I, MI, DI, TI, SI and Admix (AI).
For the most challenging case in the table, attacking the black-box IncV3 (let's take RN50 $\Rightarrow$ IncV3 attack as an example),
StyLess significantly improves the attack success rate of baseline attacks (I and MI):
46.2\% $\rightarrow$ 68.3\% (I), 59.2\% $\rightarrow$ 78.9\% (MI).
StyLess also demonstrates its capabilities when using other DNNs as the surrogate network.
For instance, DN121 $\Rightarrow$ SNv2 attack, StyLess significantly improves the baseline:
69.3\% $\rightarrow$ 91.4\% (I), 77.4\% $\rightarrow$ 95.1\% (MI).

StyLess can be combined with other attack techniques.
Previous work has shown that combining various attack methods results in powerful and transferable attacks.
StyLess can be integrated with existing combination-based attacks to enhance attack transferability.
We report the results of four combinations of existing attacks: MDI, MTDI, MTDSI, and MTDAI.
StyLess further enhances these four attack methods’ attack success rate by +8.0\%, +8.2\%, +3.0\%, and +3.3\% (in the case of RN50 $\Rightarrow$ IncV3 attack).
We evaluate attack performance using various surrogate models, including RN50, WRN101 and DN121.
For instance, when combining MTDAI with StyLess, our method enhances the transferability of MTDAI:
WRN101 $\Rightarrow$ IncV3 attack: +4.7\%, WRN101$\Rightarrow$ MNv2 attack: +1.5\%, WRN101$\Rightarrow$ SNv2 attack: +3.4\%,
DN121 $\Rightarrow$ IncV3 attack: +4.3\%, DN121 $\Rightarrow$ MNv2 attack: +0.6\%, DN121 $\Rightarrow$ SNv2 attack: +3.8\%.
The results show that StyLess is an efficient and generic approach for improving attack transferability.

\begin{table*}[ht]
    \setlength{\abovecaptionskip}{-0.2cm}
\small
\caption{Attacking unsecured black-box models with StyLess.}
\label{table:full-unsecured}
\begin{center}
        \begin{tabular}{c|c|cccccccc}
            \toprule
            Source & Attack & VGG19 & RN50 & WRN101 & DN121&IncV3&MNv2 &SNv2\\
            \midrule
\multirow{6}{*}{RN50}
& I / +Ours    & 72.8 / \textbf{88.8} & 100 / 100 & 80.8 / \textbf{97.1} & 83.0 / \textbf{97.8} & 46.2 / \textbf{68.3} & 77.1 / \textbf{91.9} & 60.6 / \textbf{77.8} \\
& MI / +Ours   & 82.9 / \textbf{94.1} & 100 / 100 & 83.9 / \textbf{97.2} & 87.5 / \textbf{98.7} & 59.2 / \textbf{78.9} & 83.8 / \textbf{93.2} & 72.3 / \textbf{83.5} \\
& MDI / +Ours  & 97.5 / \textbf{99.2} & 100 / 100 & 98.2 / \textbf{99.8} & 99.4 / \textbf{100} & 89.5 / \textbf{97.5} & 98.1 / \textbf{99.9} & 88.5 / \textbf{96.9}  \\
& MTDI / +Ours & 98.6 / \textbf{99.7} & 100 / 100 & 99.2 / \textbf{100} & 99.8 / \textbf{100} & 90.2 / \textbf{98.4} & 98.7 / \textbf{100} & 90.8 / \textbf{98.1} \\
&MTDSI / +Ours & 98.6 / \textbf{99.2} & 100 / 100 & 99.5 / \textbf{100} & 99.8 / \textbf{100} & 96.2 / \textbf{99.2} & 98.7 / \textbf{99.9} & 96.2 / \textbf{98.2}\\
&MTDAI / +Ours & 99.3 / \textbf{99.6} & 100 / 100 & 99.8 / \textbf{100} & 99.9 / \textbf{100} & 95.5 / \textbf{98.8} & 99.7 / \textbf{100} & 95.7 / \textbf{99.0} \\
\hline
\multirow{6}{*}{WRN101}
& I / +Ours     & 64.8 / \textbf{79.6} & 88.6 / \textbf{98.7} & 100 / {100} & 77.7 / \textbf{94.2} & 43.2 / \textbf{66.6} & 68.1 / \textbf{84.7} & 58.1 / \textbf{73.2} \\
& MI / +Ours    & 75.6 / \textbf{86.4} & 89.8 / \textbf{98.7} & 100 / {100} & 83.9 / \textbf{96.4} & 57.6 / \textbf{73.7} & 76.1 / \textbf{87.0} & 70.2 / \textbf{81.2} \\
& MDI / +Ours   & 92.2 / \textbf{98.5} & 99.0 / \textbf{99.9} & 100 / {100} & 97.1 / \textbf{99.9} & 86.3 / \textbf{96.3} & 93.6 / \textbf{99.0} & 86.8 / \textbf{95.8} \\
& MTDI / +Ours  & 92.5 / \textbf{98.6} & 99.2 / \textbf{100}  & 100 / {100} & 97.9 / \textbf{99.6} & 89.2 / \textbf{97.9} & 95.4 / \textbf{99.3} & 88.4 / \textbf{97.0} \\
& MTDSI / +Ours & 95.9 / \textbf{98.9} & 99.6 / \textbf{100}  & 100 / {100} & 99.7 / \textbf{100} & 95.8 / \textbf{99.5} & 97.7 / \textbf{99.9} & 94.5 / \textbf{98.4} \\
&MTDAI / +Ours  & 96.5 / \textbf{99.2} & 99.9 / \textbf{99.9} & 100 / {100} & 99.4 / \textbf{100} & 93.7 / \textbf{98.4} & 98.1 / \textbf{99.6} & 94.6 / \textbf{98.0} \\
\hline
\multirow{6}{*}{DN121}
& I / +Ours     & 79.7 / \textbf{97.7} & 87.0 / \textbf{99.4} & 74.7 / \textbf{98.0} & 100 / {100} & 55.7 / \textbf{88.7} & 81.3 / \textbf{97.0} & 69.3 / \textbf{91.4} \\
& MI / +Ours    & 86.9 / \textbf{99.1} & 89.4 / \textbf{99.7} & 78.2 / \textbf{98.6} & 100 / {100} & 66.0 / \textbf{95.7} & 86.5 / \textbf{98.7} & 77.4 / \textbf{95.1} \\
& MDI / +Ours   & 96.7 / \textbf{99.9} & 98.9 / \textbf{100} & 95.4 / \textbf{99.8}  & 100 / {100} & 90.5 / \textbf{99.0} & 97.5 / \textbf{100} & 90.9 / \textbf{98.6} \\
& MTDI / +Ours  & 97.5 / \textbf{99.9} & 99.1 / \textbf{100} & 95.9 / \textbf{99.8}  & 100 / {100} & 92.7 / \textbf{99.4} & 96.9 / \textbf{100} & 92.2 / \textbf{98.6} \\
& MTDSI / +Ours & 97.4 / \textbf{99.9} & 99.3 / \textbf{100} & 98.0 / \textbf{99.8}  & 100 / {100} & 96.5 / \textbf{99.8} & 98.9 / \textbf{100} & 95.1 / \textbf{99.1} \\
&MTDAI / +Ours  & 99.2 / \textbf{99.9} & 99.8 / \textbf{100} & 98.1 / \textbf{99.9}  & 100 / {100} & 95.1 / \textbf{99.4} & 99.4 / \textbf{100} & 95.2 / \textbf{99.0} \\
            \bottomrule
   \end{tabular}
\end{center}
\end{table*}

\subsection{Attacking Secured Models}\label{exp:secured}
We evaluate StyLess on three widely-used secured models: IncV3$_{\text {ens3 }}$, IncV3$_{\text {ens4 }}$ and IncResV2$_{\text {ens }}$, as shown in Table~\ref{table:defense}.
We present the results of I, MI, MDI, MTDI, and MTDSI on the three secured models using two surrogate models: RN50 and WRN101.
These secured networks are more robust than the unsecured DNNs we mentioned above.
Taking RN50 $\Rightarrow$ IncV3$_{\text {ens3 }}$ attack as an example,
I, MI, MDI, and MTDI achieve attack success rates of only 21.6\%, 31.5\%, 59.6\%, and 69.2\%, respectively.
StyLess effectively improves the performance of these attacks.
Notably, MTDSI has been improved by our method: 91.6\% $\rightarrow$ 97.5\% in WRN101$\Rightarrow$ IncV3$_{\text {ens3 }}$ attack;
88.6\% $\rightarrow$ 96.1\% in WRN101$\Rightarrow$ IncV3$_{\text {ens4 }}$ attack;
83.0\% $\rightarrow$ 92.5\% in WRN101$\Rightarrow$ IncResV2$_{\text {ens }}$ attack.
StyLess demonstrates its great power to enhance transferability in breaking these challenging secured DNNs.

\begin{figure}[htb]
\vspace{-2ex}
\centering
\includegraphics[width=0.45\textwidth]{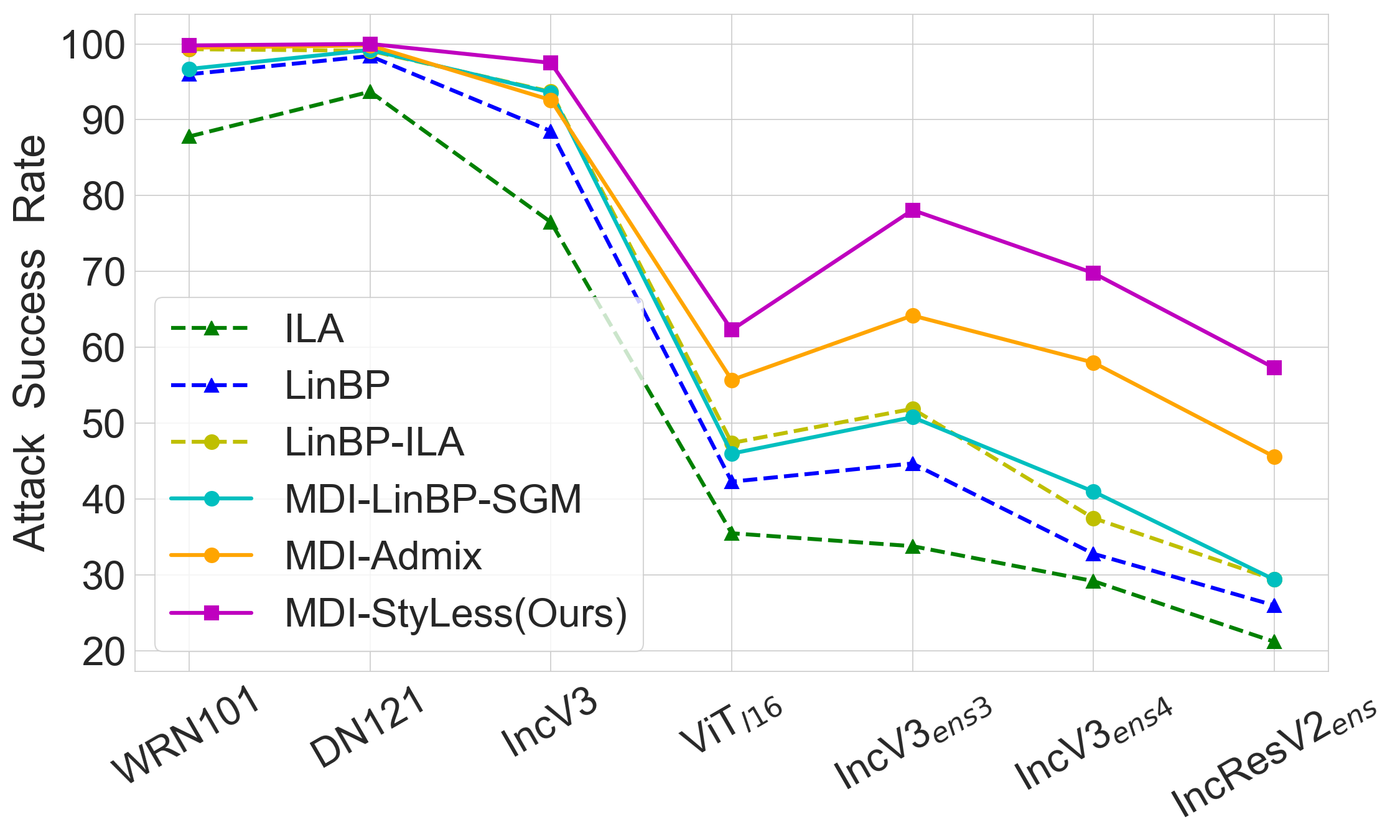}
\caption{Comparing with LinBP on different black-box models and defenses.
The surrogate model is RN50.
}
\label{fig:other_attack_on_defense}
\vspace{-3ex}
\end{figure}

Figure~\ref{fig:other_attack_on_defense} shows a comparison of StyLess and LinBP.
We consider LinBP and its combination with existing attacks such as ILA, SGM, and MDI on various target DNNs.
Our method exhibits the best attack transferability.

\begin{table}[tb]\small
    \setlength{\abovecaptionskip}{0.2cm}
    \setlength{\belowcaptionskip}{-0.4cm}
\centering
\caption{Attacking three secured black-box models with StyLess. The surrogate model is RN50 or WRN101.}
\label{table:defense}
\begin{tabular}{c|ccc}
        \toprule
        \multirow{2}{*}{Attack} & \multicolumn{3}{c}{\underline{\textbf{RN50}} $\Rightarrow$}\\
                               & IncV3$_{\text {ens3 }}$ & IncV3$_{\text {ens4 }}$ & IncResV2$_{\text {ens }}$ \\
        \midrule 
        I / +Ours    & 21.6 / \textbf{34.6} & 18.9 / \textbf{32.0} & 14.1 / \textbf{21.5} \\
        MI/ +Ours    & 31.5 / \textbf{47.1} & 29.3 / \textbf{42.4} & 20.8 / \textbf{31.0} \\
        MDI / +Ours  & 59.6 / \textbf{78.1} & 53.5 / \textbf{69.8} & 38.3 / \textbf{57.3} \\
        MTDI / +Ours & 69.2 / \textbf{89.6} & 63.8 / \textbf{81.8} & 54.6 / \textbf{72.2} \\
        MTDSI / +Ours& 88.0 / \textbf{93.1} & 84.7 / \textbf{91.3} & 77.8 / \textbf{84.5} \\
        \midrule
        & \multicolumn{3}{c}{\underline{\textbf{WRN101}} $\Rightarrow$}\\
        \midrule 
        I / +Ours    & 23.4 / \textbf{40.3} & 20.9 / \textbf{34.6} & 15.5 / \textbf{25.6} \\
        MI/ +Ours    & 35.0 / \textbf{51.2} & 30.4 / \textbf{46.6} & 24.7 / \textbf{35.9} \\
        MDI / +Ours  & 64.0	/ \textbf{81.5} & 58.8 / \textbf{76.2} & 46.9 / \textbf{66.4} \\
        MTDI / +Ours & 75.5	/ \textbf{91.4} & 70.5 / \textbf{87.4} & 63.6 / \textbf{81.0} \\
        MTDSI / +Ours& 91.6	/ \textbf{97.5} & 88.6 / \textbf{96.1} & 83.0 / \textbf{92.5} \\
        \bottomrule
    \end{tabular}
\vspace{-2ex}
\end{table}

\begin{table*}[htb]\small
\setlength{\abovecaptionskip}{-0.2cm}
\caption{Combining with ensemble-based attack method.}
\label{table:ensemble-unsecured}
\begin{center}
        \begin{tabular}{c|c|cccccccccc}
            \toprule
            \multirow{2}{*}{ $\epsilon$} &\multirow{2}{*}{Attack} & \multicolumn{10}{c}{\underline{RN50+WRN101+DN121} $\Rightarrow$}\\
                               & & VGG19 & RN50 & WRN101 & DN121&IncV3&MNv2 &SNv2 & IncV3$_{\text {ens3 }}$ & IncV3$_{\text {ens4 }}$ & IncResV2$_{\text {ens }}$ \\
            \midrule
            \multirow{3}{*}{4}
            & MI      & 56.8 & 96.4 & 94.8 & 97.2 & 35.7 & 62.3 & 48.3 & 18.4 & 15.8 & 9.4 \\
            & MTDI    & 79.2 & 97.9 & 96.6 & 98.8 & 67.5 & 84.5 & 71.6 & 36.5 & 32.1 & 22.3 \\ 
            &+StyLess & \textbf{90.4} & \textbf{99.5} & \textbf{99.4} & \textbf{99.9} & \textbf{83.2} & \textbf{94.3} & \textbf{85.4} & \textbf{50.0} & \textbf{42.9} & \textbf{29.1} \\
            \cline{1-12}
            \multirow{3}{*}{8}
            & MI      & 79.2 & 99.6 & 98.7 & 99.4 & 58.3 & 83.3 & 68.0 & 31.4 & 26.8 & 18.2\\
            & MTDI    & 95.2 & 99.9 & 99.5 & 99.9 & 89.5 & 97.5 & 90.4 & 67.6 & 61.8 & 46.9\\
            &+StyLess&  \textbf{99.1} & \textbf{100} & \textbf{100} & \textbf{100} & \textbf{97.5} & \textbf{99.6} & \textbf{97.2} & \textbf{85.1} & \textbf{77.4} & \textbf{66.1}\\
            \cline{1-12}
            \multirow{3}{*}{16}
            & MI      & 92.3 & \textbf{100} & \textbf{100} & \textbf{100} & 81.0 & 93.2 & 85.0 & 53.0 & 46.3 & 35.0 \\
            & MTDI    & 99.2 & \textbf{100} & \textbf{100} & \textbf{100} & 97.5 & \textbf{100} & 97.9 & 92.7 & 86.4 & 80.7 \\
            &+StyLess& \textbf{100} & \textbf{100} & \textbf{100} & \textbf{100} & \textbf{99.8} & \textbf{100} & \textbf{100} & \textbf{98.5} & \textbf{97.5} & \textbf{94.3} \\
            \bottomrule
        \end{tabular}
    \end{center}
\vspace{-3ex}
\end{table*}

\subsection{Combining with Ensemble-Based Method}
Generating adversarial examples based on multiple surrogate networks simultaneously can improve attack performance in practice \cite{LiuCLS17}.
This ensemble-based method has been shown to be a powerful attack, and StyLess can further improve it.
We use an ensemble of RN50, WRN101, and DN121 as the integrated surrogate model.

We compared our StyLess with ensemble-based MI and MTDI in Table~\ref{table:ensemble-unsecured}.
The considerable strength of the ensemble-based method can be seen when comparing with the results in Table~\ref{table:full-unsecured} and ~\ref{table:defense}.
For example, in RN50 $\Rightarrow$ IncV3 attack, the baseline attack MI only get 59.2\%, while it get
81.0\% attack success rate in the case of RN50+WRN101+DN121 $\Rightarrow$ IncV3 attack.
Noted that we generally use $\epsilon=16$, which is a standard setting.
As we can see, when $\epsilon=16$, the ensemble-based MTDI achieves an average attack success rate of
less than 90\% on the three secured networks: IncV3$_{\text {ens3 }}$, IncV3$_{\text {ens4 }}$, and IncResV2$_{\text {ens }}$.
StyLess can further boost the transferability of ensemble-based MTDI:
92.7\% $\rightarrow$ 98.5\% for IncV3$_{\text {ens3 }}$,
86.4\% $\rightarrow$ 97.5\% for IncV3$_{\text {ens4 }}$,
80.7\% $\rightarrow$ 94.3\% for IncResV2$_{\text {ens }}$.
These results show that StyLess is a different type of attack, and can work perfectly with the ensemble-based method.

We also report the experimental results with different $\epsilon$.
We use $\epsilon=4 \text{ or } 8$ to increase the difficulty to attack.
For instance, in the case of ensemble-based MTDI$\Rightarrow$ IncV3$_{\text {ens3 }}$ attack, attack success rate drops from 92.7\%
to 67.6\% when $\epsilon=8$ instead of $\epsilon=16$,
and StyLess can help ensemble-based MTDI gains +17.5\% (67.6\% $\rightarrow$ 85.1\%), which is a huge improvement.
When $\epsilon=4$, the results also demonstrate the advantages of StyLess.
This shows that StyLess consistently delivers strong transferability when faced with a more robust network that is difficult to attack.

\vspace{-1ex}
\subsection{Attacking the Google Cloud Vision API}
We use the Google Cloud Vision API as an example of real-world applications to evaluate the transferability of attacks.
This API allows us to
use
various vision features, such as image labeling and face detection. As a black-box model, regular users like us
cannot access its network architecture, training data, or defense mechanism. In this section, we will focus on attacking its image labeling feature.

To utilize the image labeling feature of the API, we need to upload images and obtain the predicted labels.
We use 1000 images from ImageNet as the target images, as described in the experimental setup.
Due to the fact that the API does not support all 1000 categories in ImageNet, our objective is to mislead the API's original top-1 prediction.

We use ResNet50 as the surrogate network to compare the baseline method MTDSI with our method.
Experimental results show that MTDSI achieves an attack success rate of 75.6\% on the Google Cloud Vision API. Our
approach, MTDSI-SyLess, achieves an attack success rate of 85.2\%, which is a 9.6\% improvement over the baseline.
Our results demonstrate that transfer-based black-box attacks pose a severe threat to real-world applications, and
StyLess can effectively boost attack transferability.

\subsection{Ablation Study}\label{exp:ablation}
In this session, we will present four ablation studies:
1) The position of the inserted IN layer;
2) The number of the generated stylized models;
3) The clean losses of stylized models and attack transferability;
4) The most important statistic of style features.

\begin{figure}[htb]
    \vspace{-2ex}
\setlength{\abovecaptionskip}{-0.1cm} %
\centering
\includegraphics[width=0.48\textwidth]{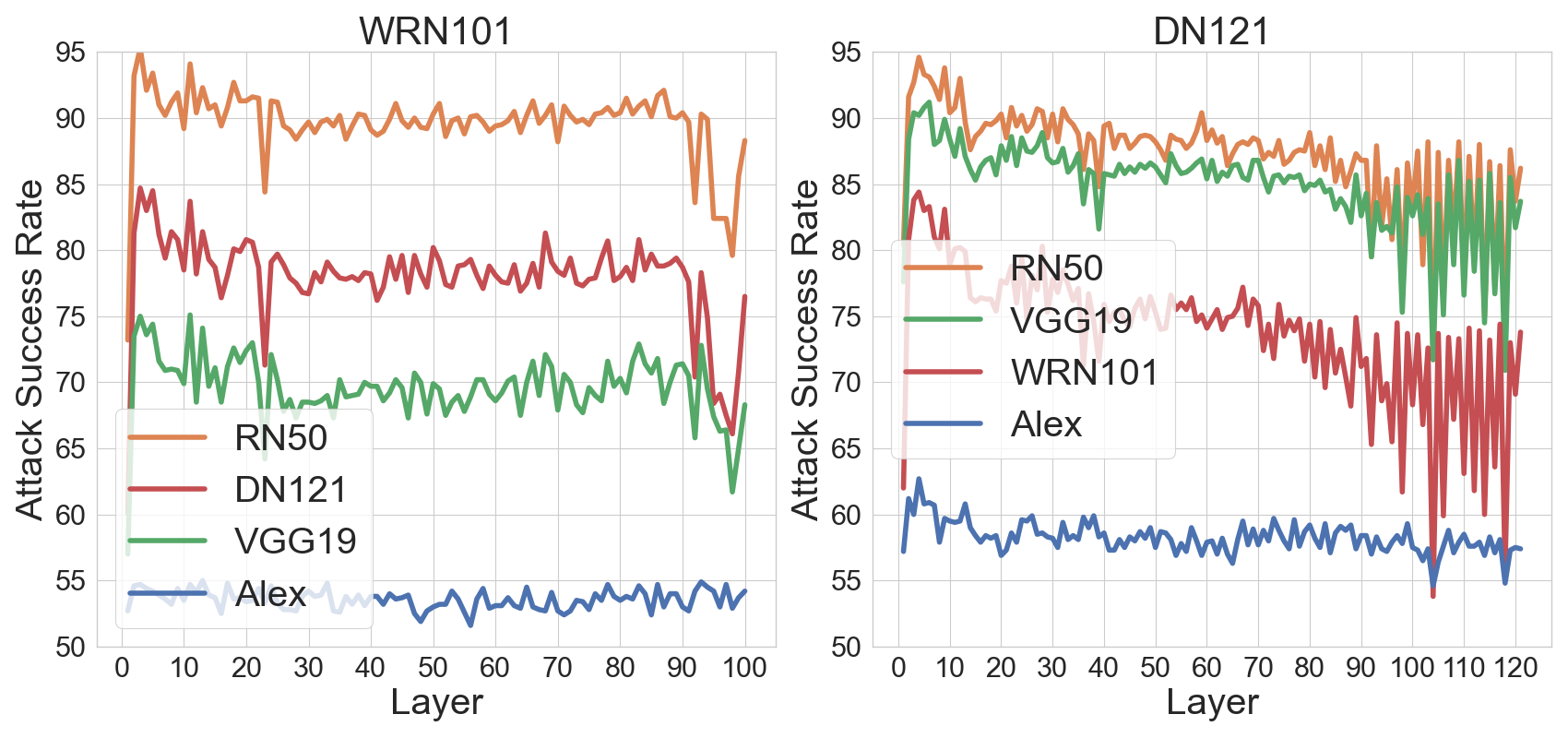}
\caption{Ablation study on using which network layer to synthesize styles. The surrogate networks are
WRN101 and DN121.}
\label{fig:ablation_layer_wrn101_dn121}
\vspace{-1ex}
\end{figure}

\textbf{Which position to insert the IN layer?}
As shown in Figure~\ref{fig:ablation_layer_wrn101_dn121},
we use features from different layers of the vanilla surrogate networks as the encoder $F_1$ for style transfer.
We evaluate the attack success rates using two surrogate networks: WRN101 and DN121.
After injecting synthesized styles into different network layers, we report the attack success rates on various DNNs,
including AlexNet, VGG19, RN50, WRN101, and DN121.
We observe a trend that the best attack success rates are usually achieved in the shallow layers of the surrogate networks.
For example, when using WRN101 as the surrogate network, injecting the synthesized styles in the layers before layer ten is a good choice.
Using intermediate layers such as layers 40 to 80 is also acceptable,
but the attack performance may be unstable or even worse when injecting the styles in the last few layers.
The results of using DN121 as the surrogate network also indicate that the last few layers are the worst choices, and the shallow layers are the optimal.

\begin{figure}[htb]
    \vspace{-1ex}
\centering
\includegraphics[width=0.48\textwidth]{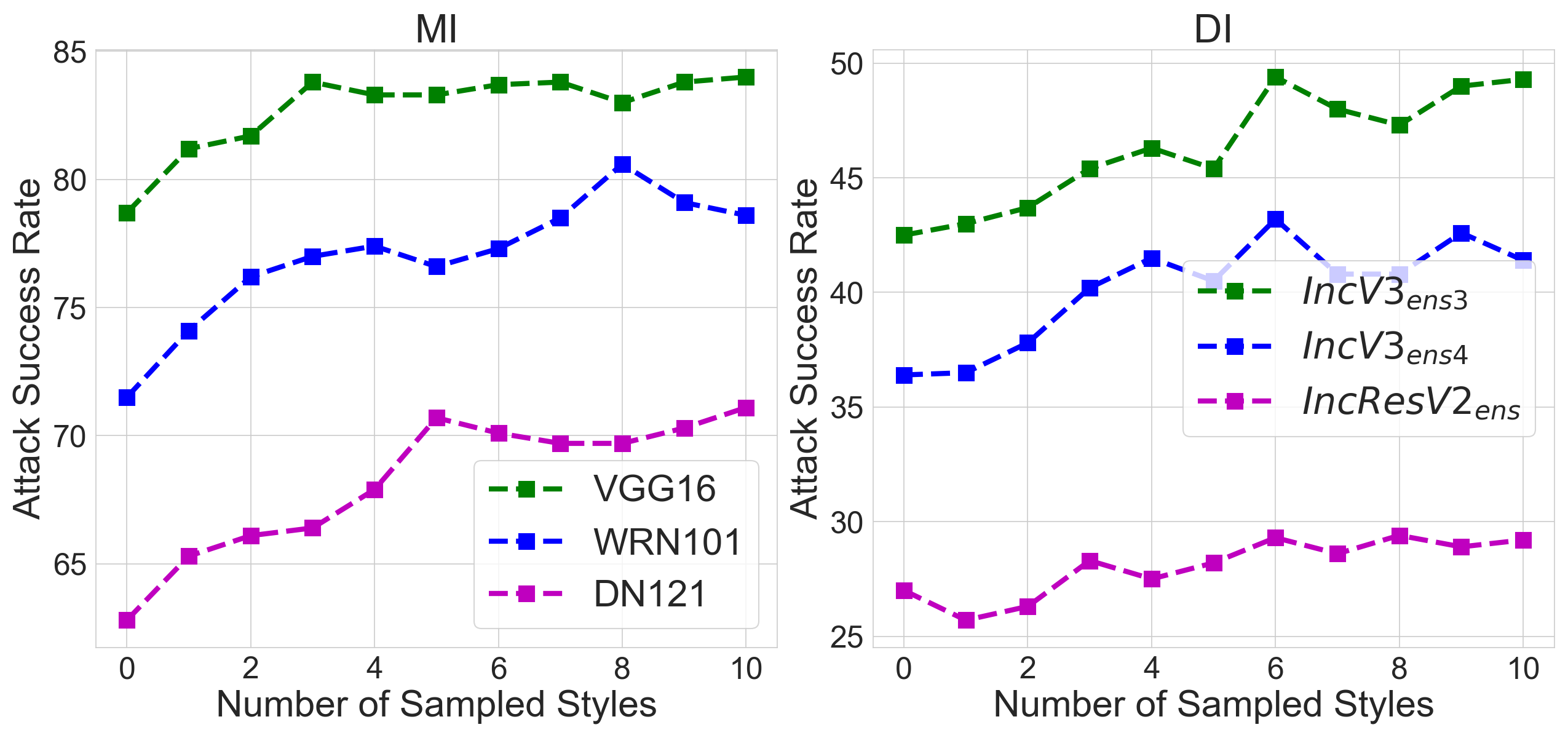}
\caption{Ablation study on the number of stylized models in an attack iteration. The surrogate network is RN50.}
\label{fig:ablation_style_num}
\vspace{-1ex}
\end{figure}

\textbf{How many stylized models should be created in each iteration?}
In Figure~\ref{fig:ablation_style_num}, we vary the number of stylized models generated in each attack iteration from zero to ten.
We conduct experiments by combining StyLess with two baseline attacks: MI and DI.
The first figure evaluates the attack success rates on three unsecured networks for our MI+StyLess attack.
In the second figure, we test DI+StyLess on three secured models.
When the number of stylized models is 0, StyLess is not involved, so it is the vanilla baseline attack.
As we can see, when the number increases from 0 to 1, the attack success rate starts to grow, which means StyLess starts to work.
There is an anomaly when DI begins to combine with StyLess.
If the number is less than three, the attack success rate on IncResV2$_{ens}$ is slightly worse than the baseline (around 25\%).
When the number increases by more than three, the attack success rate becomes higher than the baseline.
This may be because IncResV2$_{ens}$ is very strong,
and more synthesized style features need to be injected into the surrogate model.
According to the figure, StyLess works quite well when six to ten stylized models are used in an attack iteration.

\begin{figure}[htb]
\centering
\includegraphics[width=0.48\textwidth]{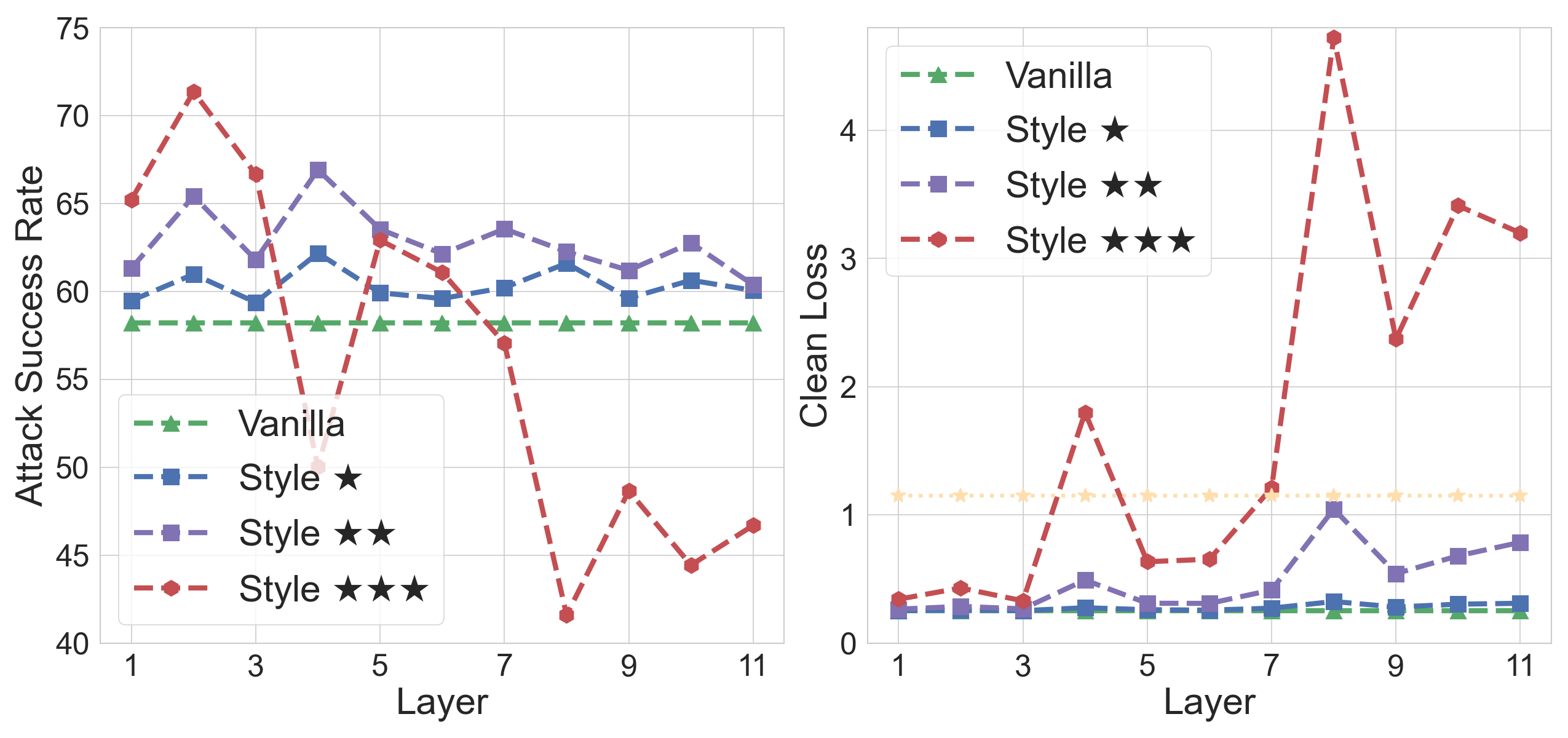}
\caption{Study how synthesized style features affect the clean loss, which in return impacts the attack success rate. The number of stars indicates the degree of change in original style features.}
\label{fig:ablation_asr_loss}
\vspace{-2ex}
\end{figure}

\textbf{How network loss affects attack success rate?}
In Figure~\ref{fig:ablation_asr_loss}, we demonstrate how varying strengths of style injection can affect network loss,
which in turn impacts attack performance on a surrogate network (in this case, VGG19).
The strength of synthesized style features is denoted by the number of stars,
with more stars indicating greater style strength that alters the style features of the surrogate model.
Generally, injecting synthesized style features should not significantly affect the clean loss,
as an increase in network loss typically leads to a decrease in clean accuracy.
Overly corrupted stylized surrogate models can also result in bad gradients for attack methods.
Therefore, there is an upper bound on clean loss when generating stylized models.
The red line in the figure represents the situation of overly corrupting,
in which the clean loss exceeds the estimated bound (indicated by a yellow line in the right figure), and the attack success rate drops significantly.
This demonstrates the importance of maintaining relatively good clean accuracy when creating stylized networks.
%

\begin{figure}[htb]
    \vspace{-2ex}
    \setlength{\abovecaptionskip}{-0.1cm}
    \setlength{\belowcaptionskip}{-0.1cm}
    \includegraphics[width=0.48\textwidth]{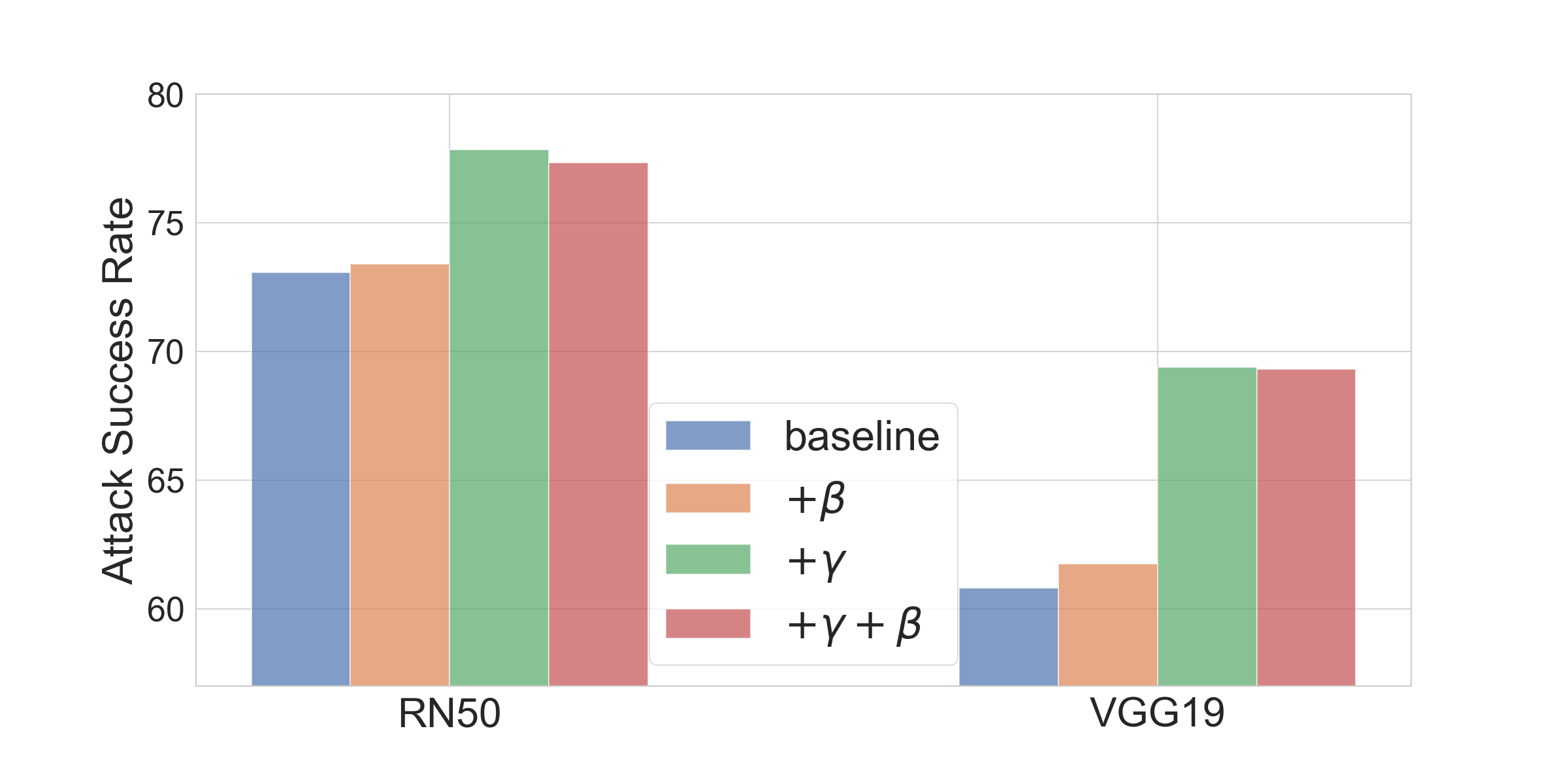}
    \caption{The effect of different statistics of style.}
    \label{fig:ablation_beta_gamma}
 \vspace{-3ex}
\end{figure}

\textbf{Which statistic of style features matters most?}
In Figure~\ref{fig:ablation_beta_gamma}, we compare the effects of the mean and variance of style features on StyLess.
Here the interpolation factor is $\lambda=0$.
Equation~\ref{eq:sip_factor} shows that $\beta$ involves the mean in IN, while $\gamma$ affects the variance.
The attack success rate represents the average success rates of attacks on five DNNs: VGG19, AlexNet, RN50, WRN101, and DN121.
The results show that $\gamma$ plays a more important role than $\beta$.
Specifically, when RN50 is used as the surrogate network, modifying $\beta$ alone barely improves the baseline attack,
while involving $\gamma$ alone enhances the attack success rate by around 5\%.
A similar observation can be made for VGG19.
From the perspective of gradient calculations, this also makes sense.
When we backpropagate through an IN layer, we have $\frac{\partial}{\partial x} \operatorname {IN} = \sigma$,
which also indicates that the variance matters most for adversarial attacks.
%

\section{Conclusion}\label{sec:conclusion}
In this work, we analyze the mechanism of attack transferability in terms of style features.
We demonstrate that existing attack methods increasingly use the style features of surrogate models during the iterative optimization, which hampers attack transferability.
To address this issue, we propose a novel attack method called StyLess to enhance transferability by reducing reliance on original style features.
StyLess uses stylized surrogate models instead of a vanilla surrogate model.
Experimental results show that StyLess outperforms existing attacks by a large margin,
and can be combined with other attack methods.
Notably, StyLess is a different paradigm from previous transferable attack methods, and we hope it will shed light on the interpretation of adversarial attacks in the future.

{\small
\bibliographystyle{ieee_fullname}
\bibliography{sample-base}
}

\end{document}